\documentclass[letterpaper, 10 pt, conference]{ieeeconf}  

\usepackage{amsmath,amssymb,amsfonts}
\usepackage{algorithm}
\usepackage{array}
\usepackage[caption=false,font=small,labelfont=rm,textfont=rm]{subfig}

\usepackage{textcomp}
\usepackage{stfloats}
\usepackage{url}
\usepackage{verbatim}
\usepackage{graphicx}
\usepackage{cite}

\usepackage{epsfig} 
\usepackage{hyperref}
\usepackage{multirow}
\usepackage{booktabs}
\usepackage{threeparttable}
\usepackage{bm}
\usepackage{algpseudocode}
\usepackage{multicol}
\usepackage[table,xcdraw]{xcolor}
\usepackage{tikz}
\usepackage{textcomp}

\IEEEoverridecommandlockouts                              

\title{\LARGE \bf
Reduce Lap Time for Autonomous Racing with Curvature-Integrated MPCC Local Trajectory Planning Method*}
\author{Zhouheng Li$^{1}$, Lei Xie$^{1, \dagger}$, Cheng Hu$^{1}$, Hongye Su$^{1}$
\thanks{* This work was supported by the Ningbo Key research and development Plan (No.2023Z116).}
\thanks{$\dagger$ Corresponding author.}
\thanks{$^{1}$ Zhouheng Li, Lei Xie, Cheng Hu, and Hongye Su are with 
	 the State Key Laboratory of Industrial, Zhejiang University, Hangzhou 310027, China. 
        {\tt\small \{ zh.li@zju.edu.cn;leix@iipc.zju.edu.cn; 22032081@zju.edu.cn;
        	hysu@iipc.zju.edu.cn\}  }  }%
}

\begin{document}

\maketitle
\thispagestyle{empty}
\pagestyle{empty}

\begin{abstract}
The widespread application of autonomous driving technology has significantly advanced the field of autonomous racing. Model Predictive Contouring Control (MPCC) is a highly effective local trajectory planning method for autonomous racing. However, the traditional MPCC method struggles with racetracks that have significant curvature changes, limiting the performance of the vehicle during autonomous racing. To address this issue, we propose a curvature-integrated MPCC (CiMPCC) local trajectory planning method for autonomous racing. This method optimizes the velocity of the local trajectory based on the curvature of the racetrack centerline. The specific implementation involves mapping the curvature of the racetrack centerline to a reference velocity profile, which is then incorporated into the cost function for optimizing the velocity of the local trajectory. This reference velocity profile is created by normalizing and mapping the curvature of the racetrack centerline, thereby ensuring efficient and performance-oriented local trajectory planning in racetracks with significant curvature. The proposed CiMPCC method has been experimented on a self-built 1:10 scale F1TENTH racing vehicle deployed with ROS platform. The experimental results demonstrate that the proposed method achieves outstanding results on a challenging racetrack with sharp curvature, improving the overall lap time by \textbf{11.4\%-12.5\%} compared to other autonomous racing trajectory planning methods. Our code is available at \textcolor{blue}{https://github.com/zhouhengli/CiMPCC}.
\end{abstract}

\section{INTRODUCTION}\label{Introduction}

\textcolor{black}{Autonomous racing requires vehicles to complete laps as fast as possible without colliding with obstacles. This mission is challenging yet crucial as it pushes the boundaries of autonomous vehicle capabilities. Achieving this mission poses significant challenges to the trajectory planning method for autonomous racing\cite{betz2022autonomous}.} \textcolor{black}{The trajectory planning methods for autonomous racing can be} divided into offline global\cite{heilmeier2019minimum,kapaniaSequentialTwoStepAlgorithm2016} and online local\cite{nilssonManoeuvreGenerationControl2014,reiterHierarchicalApproachStrategic2023,chuAutonomousHighspeedOvertaking2023} planning methods. The \textcolor{black}{offline global} trajectory planning methods can synthesize global racetrack information to generate globally optimal trajectories. However, it is challenging to account for changes in real-time vehicle states, which can pose risks to vehicle motion\cite{picottiNonlinearModelPredictiveContouring2023}. \textcolor{black}{The online local trajectory planning methods are used to tackle this problem.}  \textcolor{black}{It} can respond excellently to changing vehicle states. \textcolor{black}{The model predictive contouring control (MPCC)\cite{linigerOptimizationbasedAutonomousRacing2015} is one of the classic methods.} However, the limited information obtained makes it easy to fall into the local optimal solution\cite{chuAutonomousHighspeedOvertaking2023}. Specifically, the traditional MPCC method do not consider the effect of significant changes in the curvature of the racetrack centerline on the high-speed motion\cite{lyonsCurvatureAwareModelPredictive2023}, affecting lap time of autonomous racing. \textcolor{black}{ We tackle this challenge by integrating the curvature of the racetrack centerline into the optimization problem, optimizing the velocity profile within the planned local trajectory. This proposed method decreases the vehicle velocity in sharp curvature sections of the racetrack to keep the vehicle close to the centerline while maintaining higher velocity in sections with smaller curvature, thereby improving overall lap time.}

\begin{figure}[!t]
	\centering
	\includegraphics[scale=0.45]{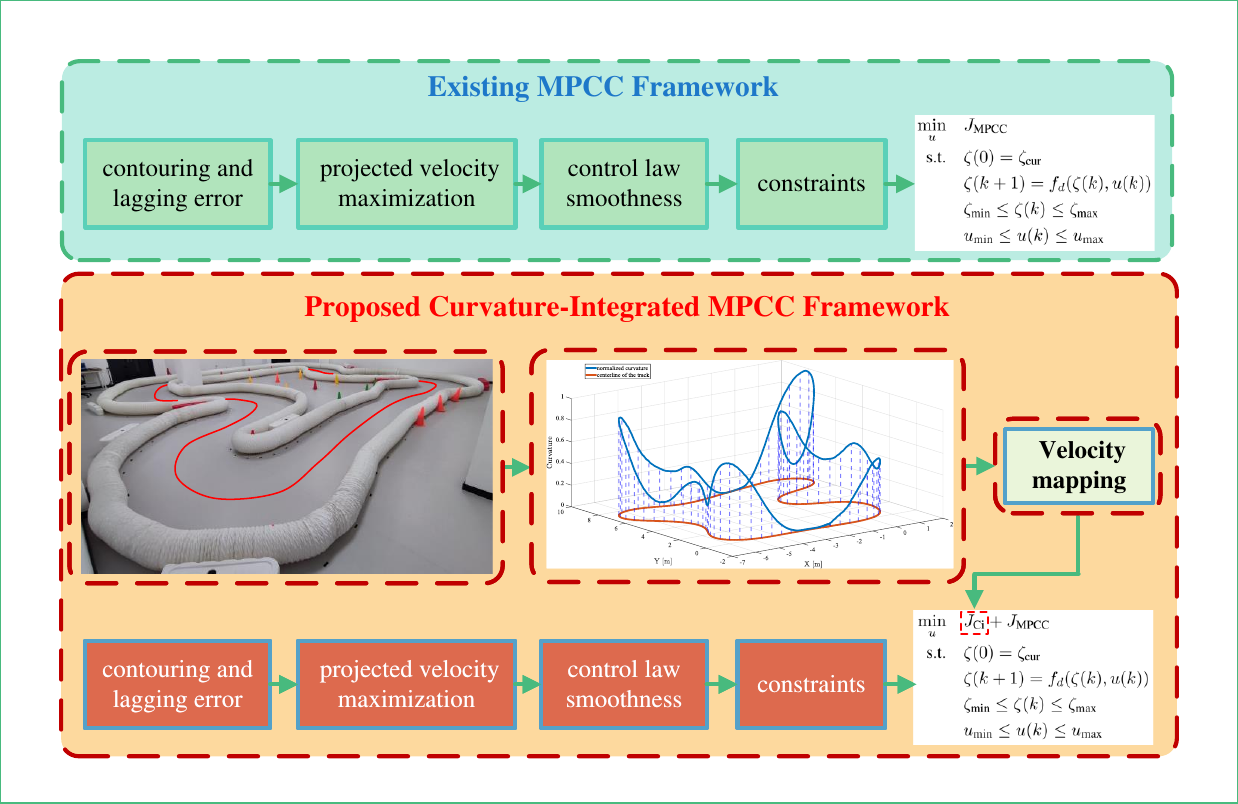}
	\caption{\textcolor{black}{The curvature-integrated MPCC (CiMPCC) trajectory planning method is illustrated in comparison with the traditional MPCC method. The CiMPCC method maps the curvature of the racetrack centerline into the optimization problem to optimize the velocity of the planned local trajectory, thereby reducing  lap time.}}
	\label{fig:teaser}
\end{figure}

In this paper, we propose a curvature-integrated MPCC (CiMPCC) trajectory planning method for autonomous racing, which tackles racetracks with significant curvature changes and performs excellent lap time. This paper has the following innovations and contributions:
\begin{enumerate}
	
	\item \textbf{\textcolor{black}{A practical method for integrating the curvature into optimization problems}}: \textcolor{black}{the curvature of the racetrack centerline is smoothed and normalized to integrate into the optimization problem of the CiMPCC method. It is utilized to optimize the velocity of the planned local trajectory, particularly in sharp and changing curvature sections of the racetrack, leading to reduced lap time};
	
	\item \textbf{A novel mapping method between curvature and velocity}: the designed mapping function maps the normalized curvature of the racetrack centerline to reference velocity. It effectively optimizes the velocity in the planned local trajectory to increase the mean velocity of a lap, thereby reducing  lap time. The CiMPCC method improves the mean velocity by \textbf{14.9\%-17.3\%} and reduces  lap time by \textbf{11.4\%-12.5\%} compared to other autonomous racing methods;
	
	\item \textbf{The well-established and reliable actual vehicle experiments:} the CiMPCC method is validated using a self-built 1:10 scale autonomous vehicle, which completes autonomous racing on a challenging racetrack.  Despite the constraints of limited computational resources and delayed execution, the proposed CiMPCC method achieves impressive mean velocity and lap time.
	
\end{enumerate}

\section{Related Works}\label{Related_Works}

This section presents commonly used methods for offline global trajectory planning and online local trajectory planning methods in autonomous racing.

The offline global trajectory planning methods for autonomous racing typically utilize path-velocity decomposition (PVD)  framework\cite{heilmeier2019minimum}. Path planning methods can be categorized based on the optimization objective into two primary types: path length-optimal\cite{braghinRaceDriverModel2008} and path curvature-optimal\cite{heilmeier2019minimum}. Velocity planning methods usually take into account the curvature changes of the planned racing path and the physical constraints of the vehicle for optimal racing performance\cite{heilmeier2020minimum}. A particularly efficient approach involves generating an initial velocity profile, which is then iteratively optimized based on the curvature of the planned path and the longitudinal force constraints of the vehicle \cite{kapaniaSequentialTwoStepAlgorithm2016}. Vehicle drifting is a frequently used strategy in racing as well. The framework proposed in\cite{weng2024aggressive} uses a mode-switching  controller to implement aggressive cornering.
	
Among the online local trajectory planning methods, MPCC is a typical representative. By introducing the concept of the contour control in industrial applications, MPCC has modeled the racing trajectory planning problem as the task of moving fastest along the racetrack centerline. Because the MPCC method does not consider the curvature of the \textcolor{black}{racetrack} centerline, the method\cite{lyonsCurvatureAwareModelPredictive2023}  improves it by introducing the curvature state into the system dynamics. However, this method is only experimented with low velocity and not extended to racing scenarios. Similar to the MPCC method, \cite{sivashangaranNonlinearModelPredictive2022} proposes an advanced nonlinear model predictive control (NMPC) strategy that uses a cost function to achieve a time-optimal trajectory by minimizing the sum of times in the prediction time horizon while maximizing the velocity. This method also does not consider the constraints imposed by the racetrack curvature on the vehicle velocity. Graph search combined with smooth spline is also an effective approach\cite{stahl2019multilayer}.

In general, offline global trajectory planning methods typically focus on planning trajectories with minimal path curvature while considering the physical constraints of the vehicle. Although these methods can generate globally optimal trajectories using comprehensive racetrack information, they lack the ability to respond to real-time vehicle states during runtime. On the other hand, MPCC, as an online local trajectory planning method, excels at dynamically responding to runtime conditions. However, it ignores the curvature of the racetrack centerline, which is crucial for optimizing autonomous racing performance. The CiMPCC method proposed in this paper fills this gap. It considers the curvature of the racetrack centerline in the optimization problem, thereby achieving performance-oriented autonomous racing.

This paper consists of several sections. Section \ref{PRELIMINARIES} describes the traditional MPCC method. Section \ref{CiMPCC} explains CiMPCC in detail, including the modeling method of the racetrack centerline and its incorporation into the optimization problem. Section \ref{EXPERIMENTAL}  presents the results of applying CiMPCC to autonomous racing with the self-built 1:10 vehicle. Section \ref{Conclusion} serves as a conclusion that summarizes the findings and outlines directions for future research.

\begin{figure}[!t]
	\centering
	\includegraphics[scale=0.26]{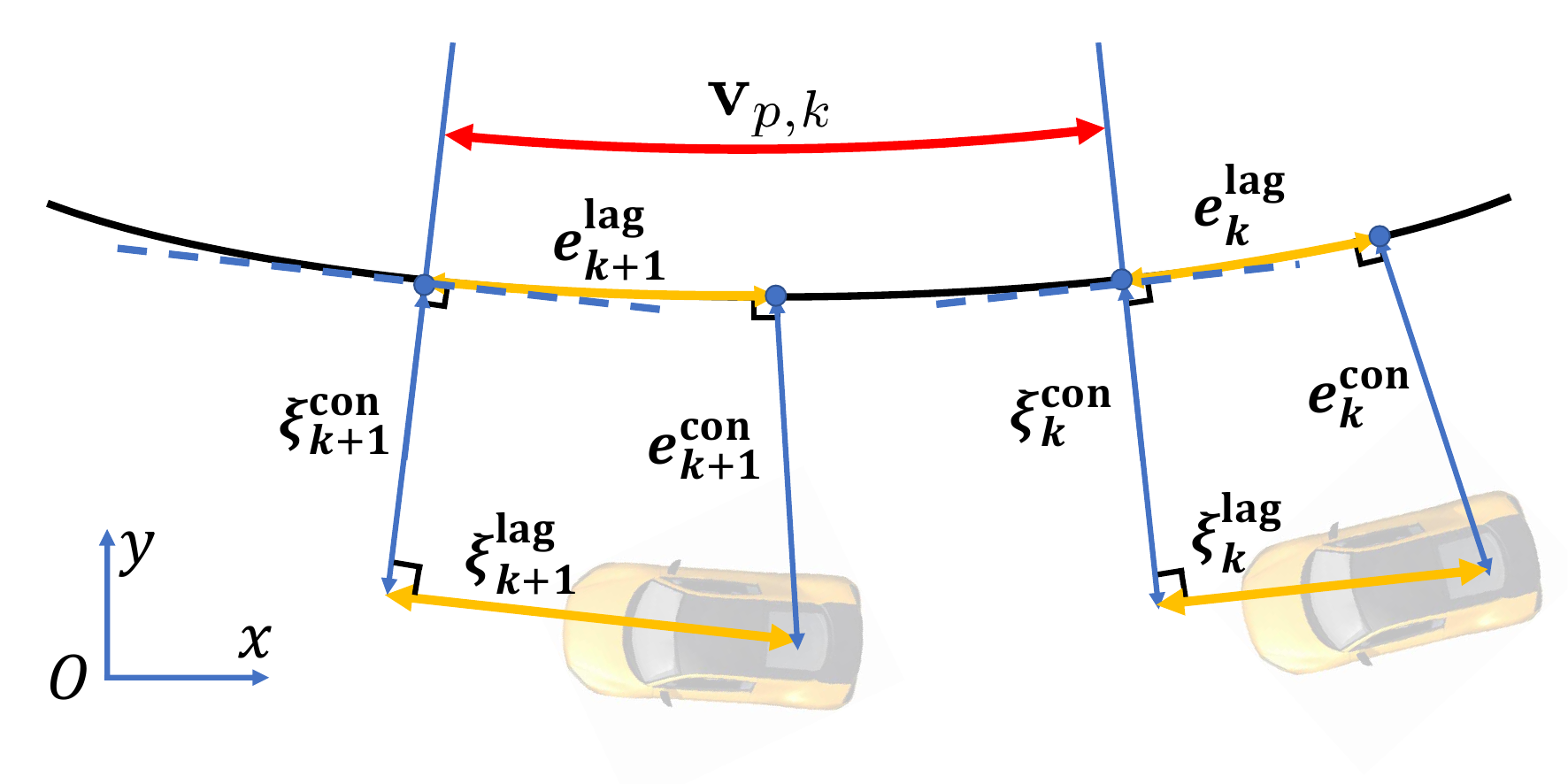}
	\caption{Schematic representation of the approximation of the contour error and lag error of MPCC.}
	\label{fig:mpcc}
\end{figure}

\section{PRELIMINARIES}\label{PRELIMINARIES}
In this section, the traditional MPCC method\cite{linigerOptimizationbasedAutonomousRacing2015} of autonomous racing trajectory planning is described. 

The differential equation of the vehicle can generally be described as: 
\begin{equation}
	\begin{aligned}\label{eq:continue}
		\dot{\zeta}=f(\zeta,u)
	\end{aligned}
\end{equation}
where $\zeta$ denotes the state of the vehicle, $u$ denotes the control space, and $f(\cdot)$ denotes the vehicle model. The commonly used vehicle kinematics models are\cite{zhangOptimizationBasedCollisionAvoidance2021,zhangTrajectoryTrackingControl2020}, and the dynamics models are\cite{zhouLearningBasedMPCController2022,qiMPCbasedControllerFramework2021,huCombinedFastControl2022}.

The~\eqref{eq:continue} can be expressed in the following form after discretization at $k$-th stage using discretization methods such as \textcolor{black}{Runge–Kutta} method\cite{zhangGuaranteedCollisionFree2021} and first-order Euler method\cite{huCombinedFastControl2022}: 
\begin{equation}
	\begin{aligned}
		\zeta(k+1)=f_d(\zeta(k),u(k))
	\end{aligned}
\end{equation}
where $\zeta(k)$ and $u(k)$ denote the discrete state and control variable of the vehicle at stage $k$ in the prediction horizon. $f_d$ is the mapping function after the discretization.

At stage $k$ in the prediction horizon, the approximate contour error, denoted as $\bm{\xi}_k^{\textrm{con}}$, represents an approximation of the actual contour error $\textbf{e}_k^{\textrm{con}}$. Similarly, the approximate lag error, denoted as $\bm{\xi}_k^{\textrm{lag}}$, serves as an approximation of the actual lag error which is represented by $\textbf{e}_k^{\textrm{lag}}$. Moreover, the projected velocity of the vehicle on the centerline of the racetrack is denoted as $\mathbf{v}_{p}$. Then the projected velocity of the vehicle at stage $k$ in the prediction horizon is represented by $\mathbf{v}_{p,k}$. The schematic is shown in Fig.~\ref{fig:mpcc}. 

Therefore, the objective function of MPCC is constructed as follows:
\begin{equation}
	\begin{split}
		J_\textrm{MPCC} = &  \sum_{k=1}^{N_p} \left( \Vert \bm{\xi}_k \Vert^2_{{Q}} - \gamma \cdot \mathbf{v}_{p,k} \cdot T_s \right) + \\
		&  \sum_{k=1}^{N_c-1} \Vert \Delta u_k \Vert^2_{{R_1}}  + \sum_{k=1}^{N_c} \Vert u_k - u_{\textrm{ref}} \Vert^2_{{R_2}}  \\
	\end{split}
\end{equation}
where $\bm{\xi}_k={\begin{bmatrix} \bm{\xi}_k^{\textrm{con}} & \bm{\xi}_k^{\textrm{lag}} \end{bmatrix}}$, $Q,\gamma$, and $R_1,R_2$ are the weight coefficients. $u_{\textrm{ref}}$ is the reference control variable. $N_p$ is the length of the prediction horizon and $N_c$ is the length of the control horizon. $T_s$ is the prediction time.

\begin{figure*}[!t]
	\centering
	\includegraphics[scale=0.84]{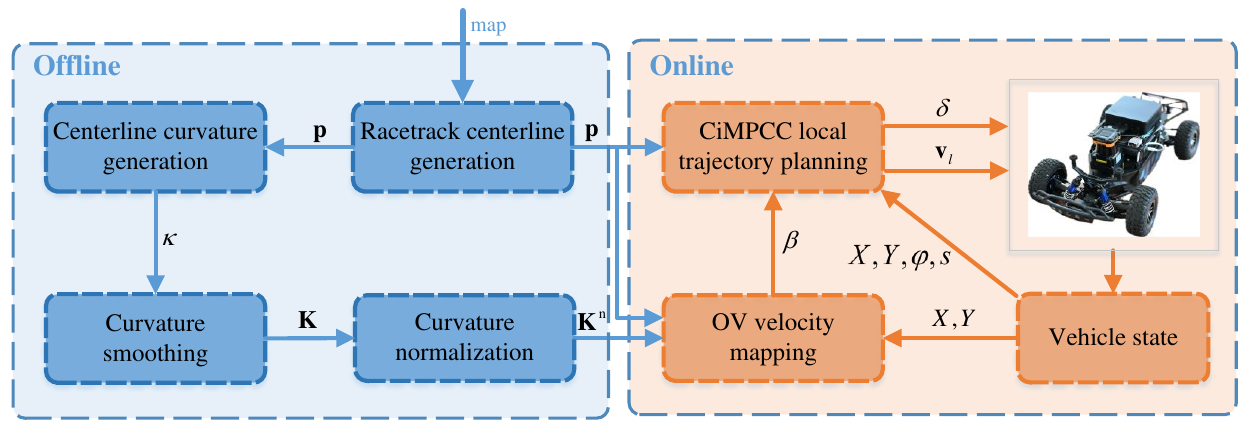}
	\caption{The schematic diagram illustrates the implementation of the CiMPCC method, which is divided into two main sections: offline and online. The offline section generates the normalized smooth curvature (NSC) of the racetrack centerline, while the online section maps the reference overall velocity (OV) based on this curvature.}
	\label{fig:total_algo}
\end{figure*}

Referring to\cite{linigerOptimizationbasedAutonomousRacing2015}, the MPCC problem can be denoted in the following formation:
\begin{subequations}
	\label{eq:mpcc}
	\begin{align}
		\min_{u} \quad & J_\textrm{MPCC} \nonumber  \\
		\text{s.t.} \quad 
		& \zeta(0) = \zeta_{\textrm{cur}} \\
		& \zeta(k+1) = f_d(\zeta(k),u(k))   \\ 
		& \zeta_{\textrm{min}} \leq \zeta(k) \leq \zeta_{\textrm{max}}  \label{cons:chg_chi}  \\
		& u_{\textrm{min}} \leq u(k) \leq u_{\textrm{max}} \label{cons:chg_u}
	\end{align}
\end{subequations}
where $\zeta_{\textrm{cur}}$ is the current state of the vehicle.

Constraints~\eqref{cons:chg_chi} and~\eqref{cons:chg_u}  represent the racetrack boundary constraints and collision avoidance constraints, as well as the constraints of the control variable. Here, $\zeta_{\textrm{max}}$ and $\zeta_{\textrm{min}}$ denote the upper and lower bounds of the state constraints, while $u_{\textrm{max}}$ and $u_{\textrm{min}}$ are the upper and lower limits of the control variable, respectively.

At this point, the introduction of the fundamentals of the MPCC method has been completed. However, as pointed out in Section~\ref{Related_Works}, the optimization problem~\eqref{eq:mpcc} does not take into account the curvature information of the racetrack centerline. Therefore, in the next section, the proposed CiMPCC method, which aims to solve this problem, is described in detail.

\section{CiMPCC Local Trajectory Planning Method}\label{CiMPCC}

\textcolor{black}{In this section, the details of the CiMPCC method are illustrated. The methods for modeling the curvature of the racetrack centerline and incorporating it into the optimization problem to optimize the velocity of the planned local trajectory are described below.}

\subsection{Curvature generation}\label{Racetrack_centerline_generation}
The racetrack centerline is first discretized to obtain a sequence of discrete points denoted as $\mathbf{p}\in \mathbb{R}^{N \times 2}$, where $N$ is the number of discrete points. The $i$-th point in $\mathbf{p}$ is denoted as $\mathbf{p}_i={\begin{bmatrix} {x}_{i} & {y}_{i} \end{bmatrix}}$. The ${x}_{i}$ and ${y}_{i}$ denote its horizontal and vertical coordinate in Cartesian frame.

The curvature $\bm{\kappa}$ corresponding to the discrete path points \textcolor{black}{$\mathbf{p}$}  on the racetrack centerline is calculated next. Take the $i$-th point $\mathbf{p}_i$ as the instance, its curvature $\bm{\kappa}_{i}$ is denoted as:
\begin{equation}
	\begin{split}
		\bm{\kappa}_{i} &=  \frac{ \Vert \Delta {{x}}_{i} \cdot 
			\Delta^2 {{y}}_{i} - \Delta^2 {{x}}_{i} \cdot \Delta {{y}}_{i} \Vert}
		{\left( \Delta {{x}}_{i}^2  + \Delta {{y}}_{i}^2 \right)^{\frac{3}{2}}} \label{eq:kappa}
	\end{split}
\end{equation}
where:
\begin{equation}
	\begin{split}
		&\Delta {{x}}_{i} = {{x}}_{i} - {{x}}_{i-1}\\
		&\Delta^2 {{x}}_{i} = {\Delta {x}}_{i} - {\Delta {x}}_{i-1}\\
	\end{split}
\end{equation}
$\Delta {{y}}_{i}$ and $\Delta^2 {{y}}_{i}$ use the same calculation method.

\subsection{Curvature smoothing and normalization}\label{Curvature_smoothing_and_normalization}
The curvature obtained after~\eqref{eq:kappa} usually has a high variation in amplitude, as shown in Fig.~\ref{fig:smooth_orig}. To ensure the stability of the reference velocity mapping, the curvature $\bm{\kappa}$ needs to be smoothed. 

Moving Average Filter (MAF)\cite{smith1997scientist} is used for curvature smoothing. For the curvature ${\bm{\kappa}}_i$, its smoothed curvature $\mathbf{K}_i$ is denoted as:
\begin{equation}
	\mathbf{K}_i = \frac{1}{\mathbf{w}} 
	\sum_{\scriptstyle m=i-\frac{\mathbf{w}-1}{2}}^{\scriptstyle i+\frac{\mathbf{w}-1}{2}} 
	\bm{\kappa}_m
\end{equation}
where $\mathbf{w}$ is the window width of the MAF. 

In order to avoid the influence of extreme curvature on the CiMPCC method, the \textcolor{black}{smoothed} curvature $\mathbf{K}$ is normalized. For the smoothed curvature $\mathbf{K}_i$, its normalized form denoted as $\mathbf{K}_i^{\textrm{n}}$ is calculated as shown below:
\begin{equation}
	\begin{split}
		\mathbf{K}_i^{\textrm{n}} &= \frac{\mathbf{K}_i - 
			\mathbf{K}_{\text{min}}}{{\mathbf{K}_{\text{max}} - 
				\mathbf{K}_{\text{min}}}}
	\end{split}
\end{equation}
where $\mathbf{K}_{\text{min}}$ and $\mathbf{K}_{\text{max}}$ are the minimum and maximum curvature in $\mathbf{K}$. 

\subsection{Velocity mapping}\label{Velocity_mapping}

The proposed CiMPCC method maps the normalized smooth curvature (NSC) to a reference velocity profile and integrates this profile into the objective function to optimize the velocity of the planned local trajectory.

\begin{figure}[!t]
	\centering
	\subfloat[Schematic representation of the smoothed curvature corresponding to the original curvature of the racetrack centerline.]{\includegraphics[scale=0.20]{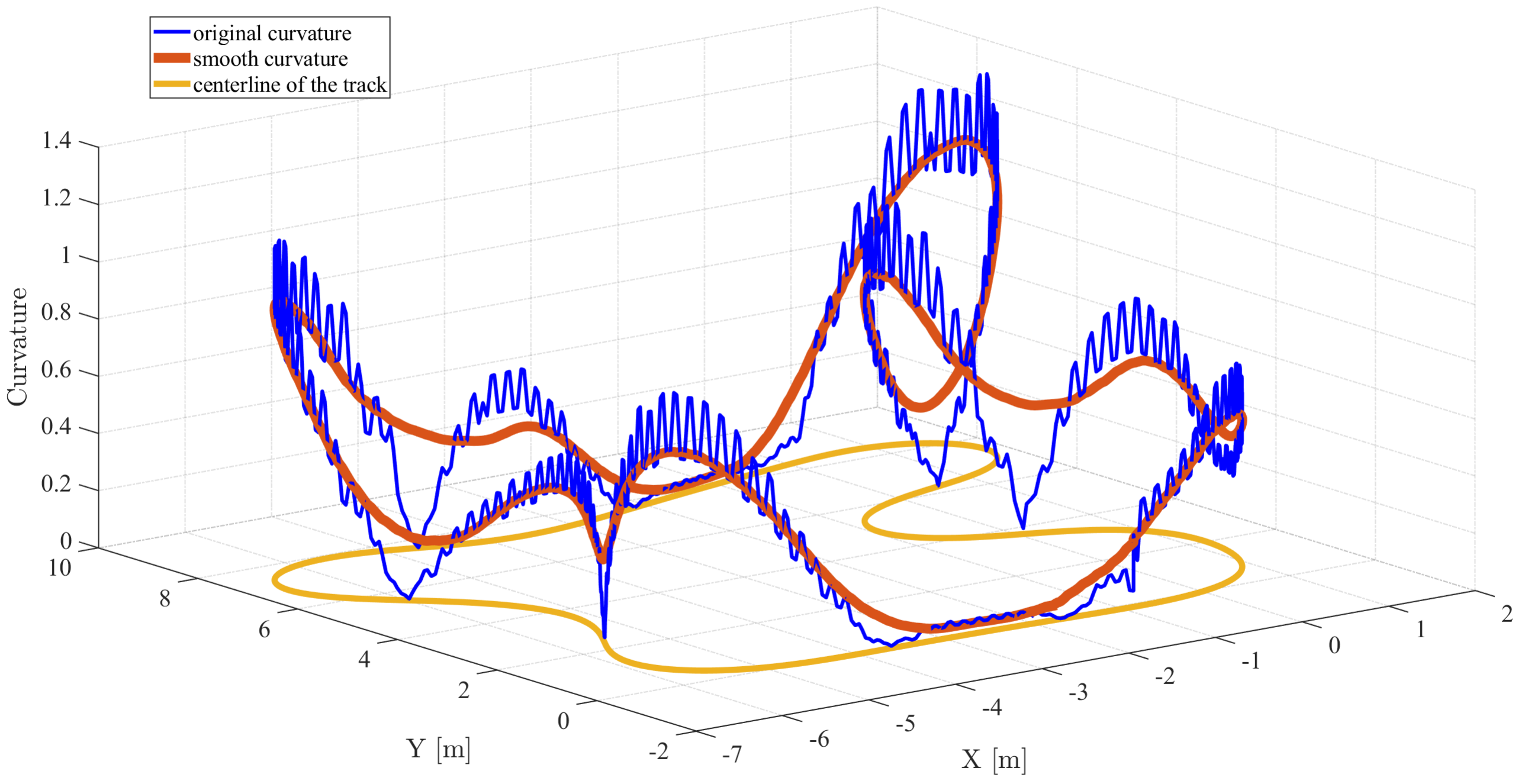}\label{fig:smooth_orig}}
	\vfil
	\subfloat[Schematic illustration of the normalized smooth curvature (NSC).]{\includegraphics[scale=0.20]{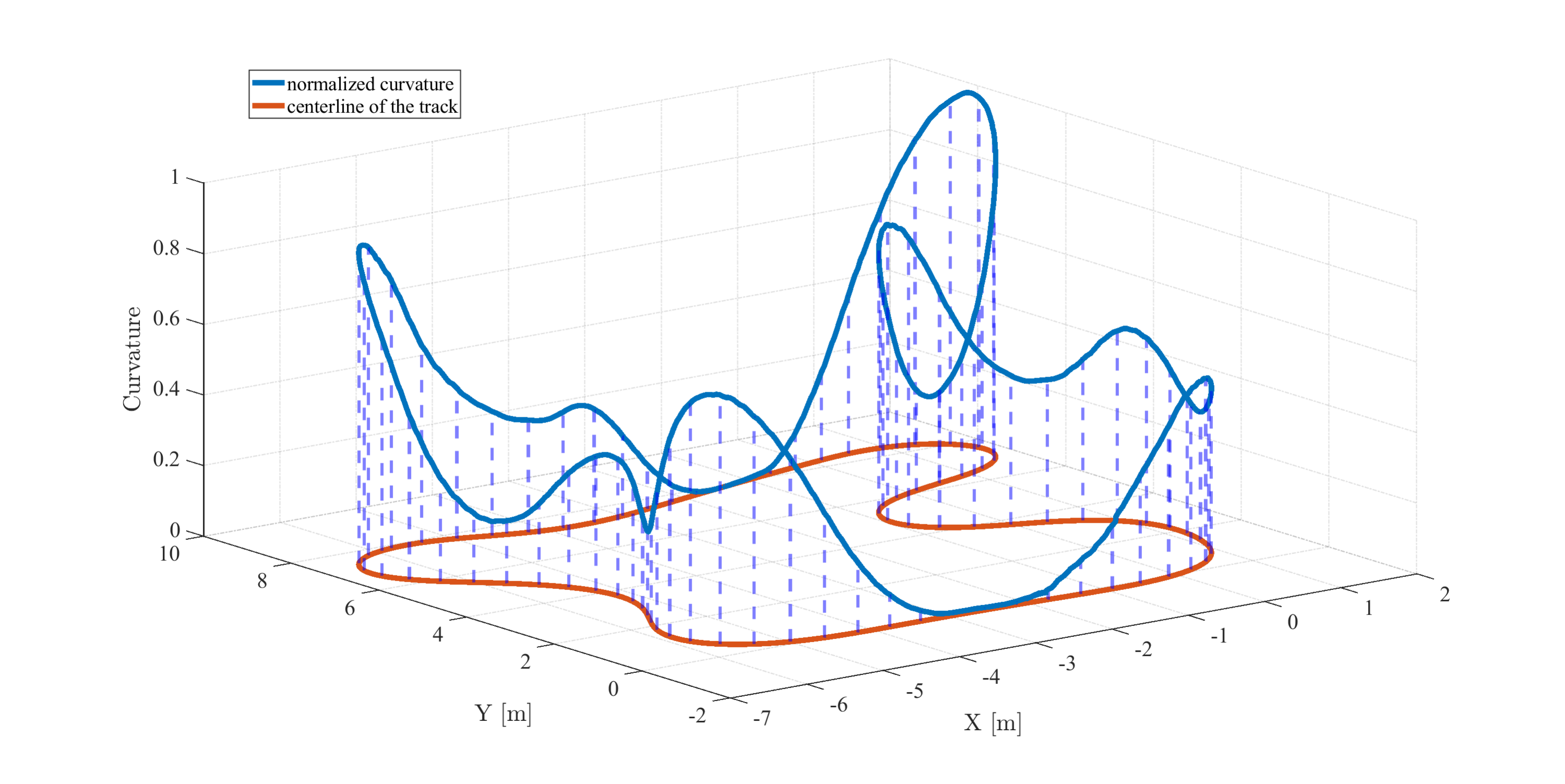}\label{fig:norm}}
	\vfil
	\subfloat[Schematic diagram of the mapping function from NSC to reference OV.]{\includegraphics[scale=0.25]{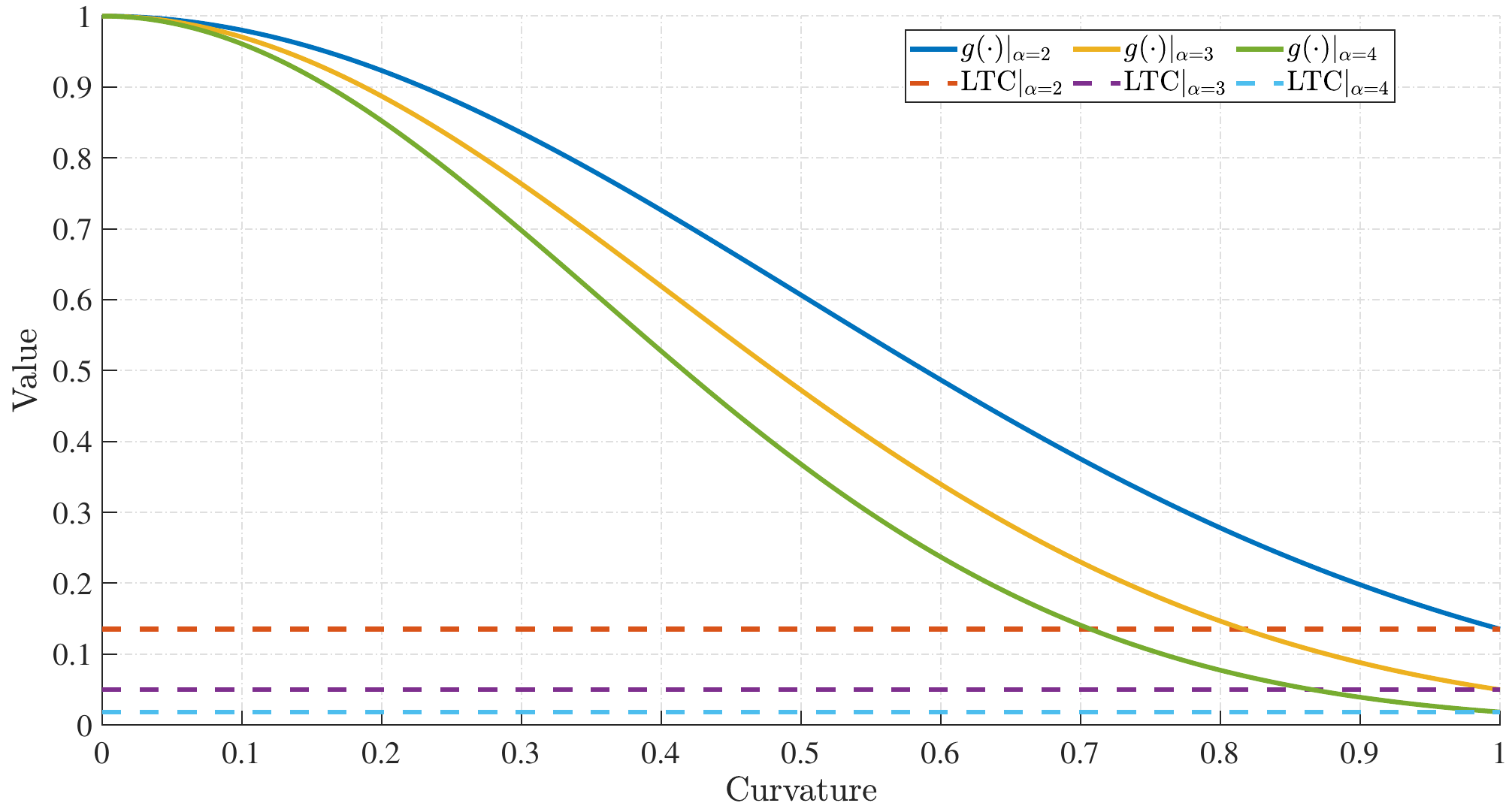}\label{fig:mapping}}
	\caption{Fig.~\ref{fig:smooth_orig} and Fig.~\ref{fig:norm} visualize the process of computing the NSC. Fig.~\ref{fig:mapping} demonstrates the NSC-velocity mapping function.}
	\label{fig:algo}
\end{figure}

Referring to\cite{dombergDeepDriftingAutonomous2022}, the mapping is represented by the following function:
\begin{equation}
	\begin{split}
		g(\mathbf{K}_i^{\textrm{n}})=e^{-\bm{\alpha}\cdot 
			(\mathbf{K}_i^{\textrm{n}})^2},\bm{\alpha}>0,i\in[1,N]
	\end{split}
\end{equation}
where $\bm{\alpha}$ is the coefficient to adjust the sensitivity of the \textcolor{black}{reference} velocity to the NSC, as shown in Fig.~\ref{fig:mapping}. Since the NSC is normalized, the mapping contains an upper truncation coefficient (UTC) $\bar{g}(0)=1$ and a lower truncation coefficient (LTC) $\underline{g}(1) = e^{-\bm{\alpha}} > 0$.

Then define the overall velocity (OV) of the vehicle  $\mathbf{v}={\begin{bmatrix} \mathbf{v}_l & \mathbf{v}_{p} \end{bmatrix}}$ and use $\mathbf{v}_k={\begin{bmatrix} \mathbf{v}_{l,k} & \mathbf{v}_{p,k} \end{bmatrix}}$ to denote the OV at the $k$-th stage in the prediction horizon. $\mathbf{v}_{l}$ is the longitudinal velocity of the vehicle. Therefore, the objective function $J_\textrm{Ci}$ incorporating NSC can be expressed as follows::
\begin{equation}
		J_\textrm{Ci} =  \sum_{k=1}^{N_p} 
		\left(1 - \bm{\beta}\right) \cdot
		\Vert \mathbf{v}_k - \underline{\mathbf{v}} \Vert^2_{{R_3}}
		+ 
		 \bm{\beta} \cdot 
		\Vert \mathbf{v}_k - \bar{\mathbf{v}} \Vert^2_{{R_3}}, 
		 \label{eq:J}
\end{equation}
where $\bm{\beta}=g({\mathbf{K}}_{\textrm{cur}}^{\textrm{n}})$, and ${\mathbf{K}}_{\textrm{cur}}^{\textrm{n}}$ denotes the NSC corresponding to the point $\mathbf{p}_{\textrm{cur}}$ that the vehicle is closest to at the current instant. The upper and lower bounds of the $\mathbf{v}$ during autonomous racing are adjustable parameters denoted by $\bar{\mathbf{v}}$  and ${\mathbf{\underline{v}}}$ respectively. They represent aggressive and safe racing velocities. $R_3$ is the weight coefficient.

\subsection{MPC formulation}\label{MPC_formulation}
The MPC formulation of the proposed CiMPCC method, along with a discussion, is demonstrated below. To begin with, the advantages of the proposed NSC-velocity mapping are analyzed. The objective function $J_\textrm{Ci}$ implies that the OV is reduced as much as possible when the curvature of the racetrack centerline is large. Because of the LTC, it is not reduced to the lowest velocity thus ensuring racing performance. On the contrary, when the curvature of the racetrack decreases, the OV of the vehicle increases rapidly, at which time the vehicle is able to approach the maximum velocity because the UTC $\bar{g}(0)=1$.

So far, the following MPC form of the CiMPCC method can be obtained based on the optimization problems~\eqref{eq:mpcc} and~\eqref{eq:J}:
\textcolor{black}{
\begin{equation}
	\begin{aligned}\label{cimpcc}
		\min_{u} \quad & J_\textrm{Ci} +  J_\textrm{MPCC} \nonumber \\
		\text{s.t.} \quad & \zeta(0) = \zeta_{\textrm{cur}} \nonumber \\
		& \zeta(k+1)=f_d(\zeta(k),u(k))  \nonumber \\ 
		& \zeta_{\textrm{min}} \leq \zeta(k) \leq \zeta_{\textrm{max}} \\
		&u_{\textrm{min}} \leq u(k) \leq u_{\textrm{max}} \nonumber \\
	\end{aligned}
\end{equation}
}

\begin{figure*}[!t]
	\centering
	\subfloat[The hardware architecture of the Driving, Drifting, and Racing Automation (DDRA) vehicle.]{\includegraphics[scale=0.27]{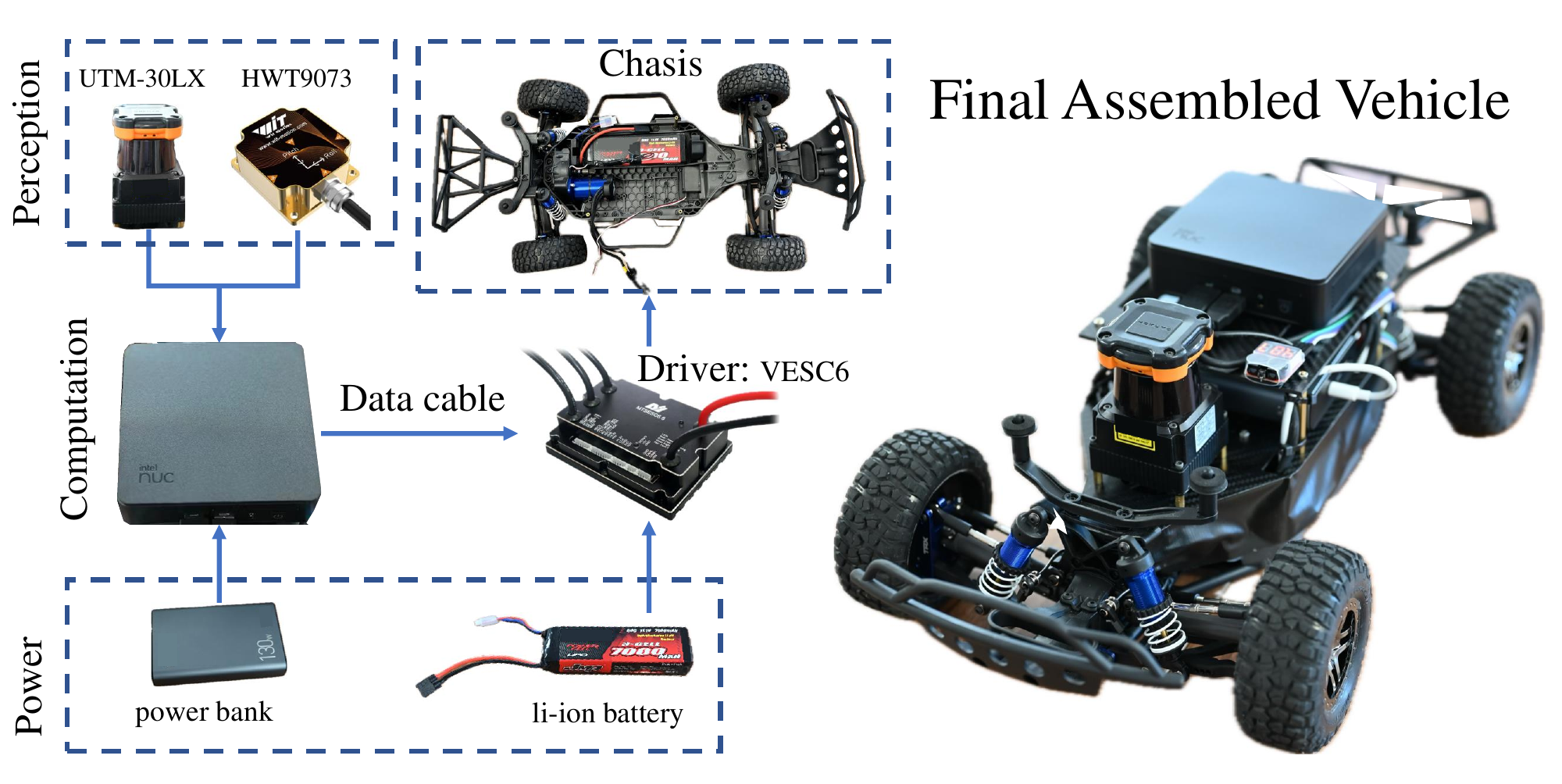}\label{fig:real_car}}
	\hfil
	\subfloat[The self-constructed racetrack with changing and sharp curvature.]{\includegraphics[scale=0.24]{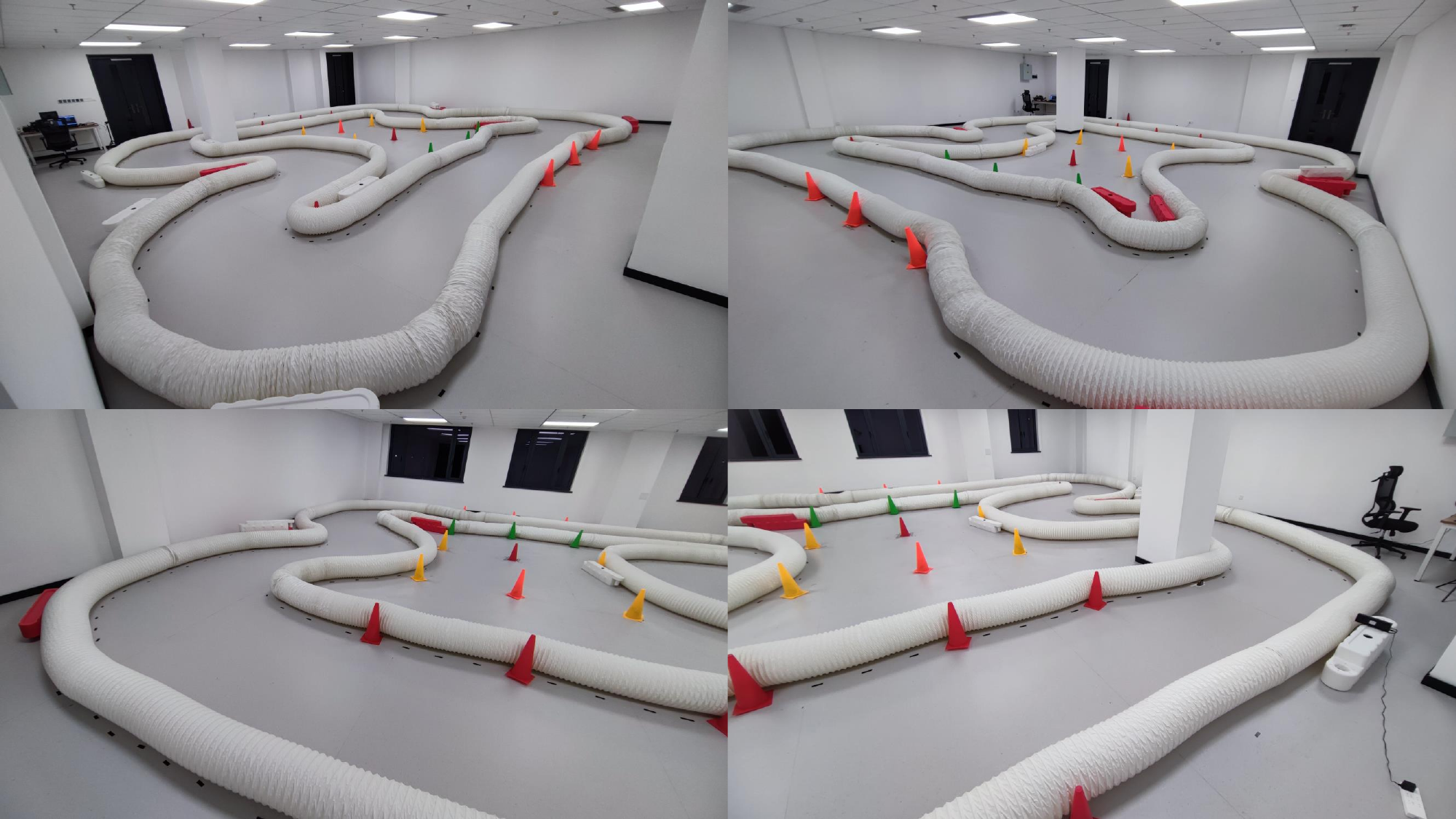}\label{fig:racetrack}}
	\hfil
	\subfloat[The schematic of the ROS architecture in the DDRA vehicle. \textcolor{black}{
		The Erpm is obtained from the VESC, the 2D lidar used is the UTM-30LX, and the IMU utilized is the HWT9073.}]{\includegraphics[scale=1.5]{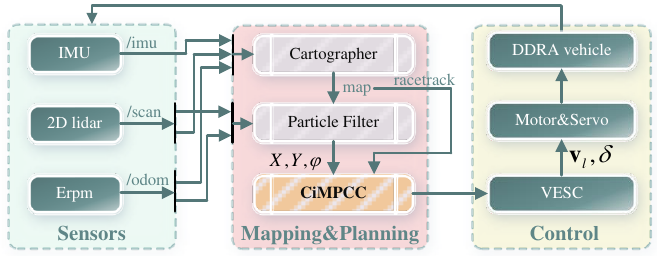}\label{fig:ros_sys}}
	\caption{\textcolor{black}{The hardware and software architecture of the DDRA vehicle are used to validate the CiMPCC method, and the racetrack setting is demonstrated.}}
	\label{fig:real_env}
\end{figure*}

By incorporating the NSC into the optimization problem, CiMPCC can optimize the velocity of the vehicle on sections of the racetrack where the curvature changes significantly. Simultaneously, due to the excellent continuity of the introduced mapping $g(\cdot)$, the planned velocity profile corresponding to the planned trajectory of CiMPCC maintains a high level of continuity.

The implementation of CiMPCC is described in detail in this section. The proposed method first smooths the curvature of the racetrack centerline using MAF and then obtains the NSC after normalization. Finally, the NSC is molded into the optimization problem through  a continuous function. The effectiveness of the CiMPCC method is verified in the next section.

\section{EXPERIMENTAL RESULTS AND DISCUSSION}\label{EXPERIMENTAL}

In this section, we analyze the performance of the CiMPCC method as applied to a physical vehicle and compare it with other autonomous racing methods. We verify the effectiveness of the CiMPCC method by analyzing velocity, lap time, and computation time.

\begin{figure*}[!t]
	\centering
	\subfloat[The vehicle velocity $\mathbf{v}_l$ changes with the projection point. \textcolor{black}{Compared to the MPCC method, the CiMPCC method maintains higher vehicle velocities on racetrack sections with small curvature.}]{\includegraphics[scale=0.26]{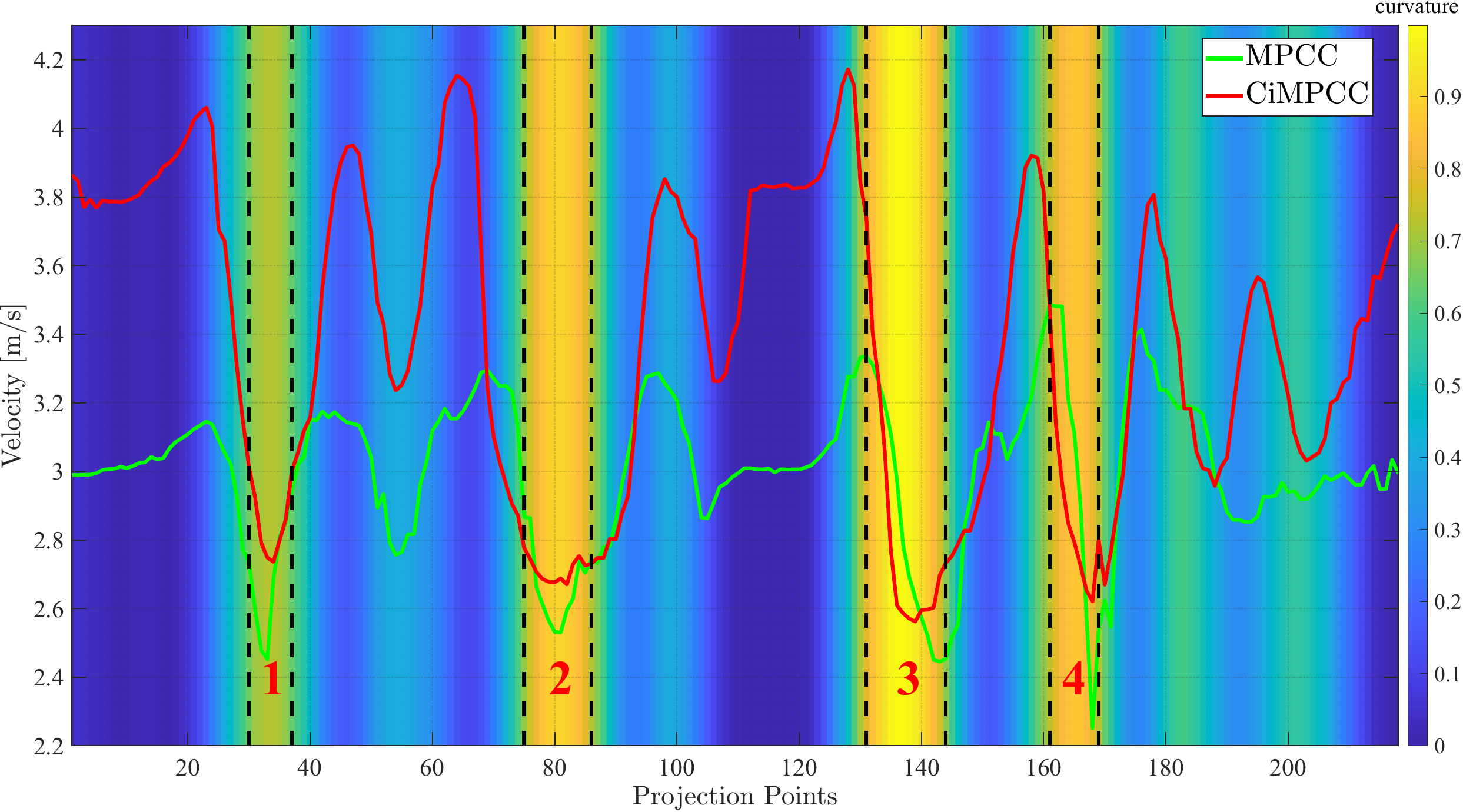}\label{fig:samples}}
	\hfil
	\subfloat[The map obtained by the cartographer and the corresponding racetrack centerline.]{\includegraphics[scale=0.27]{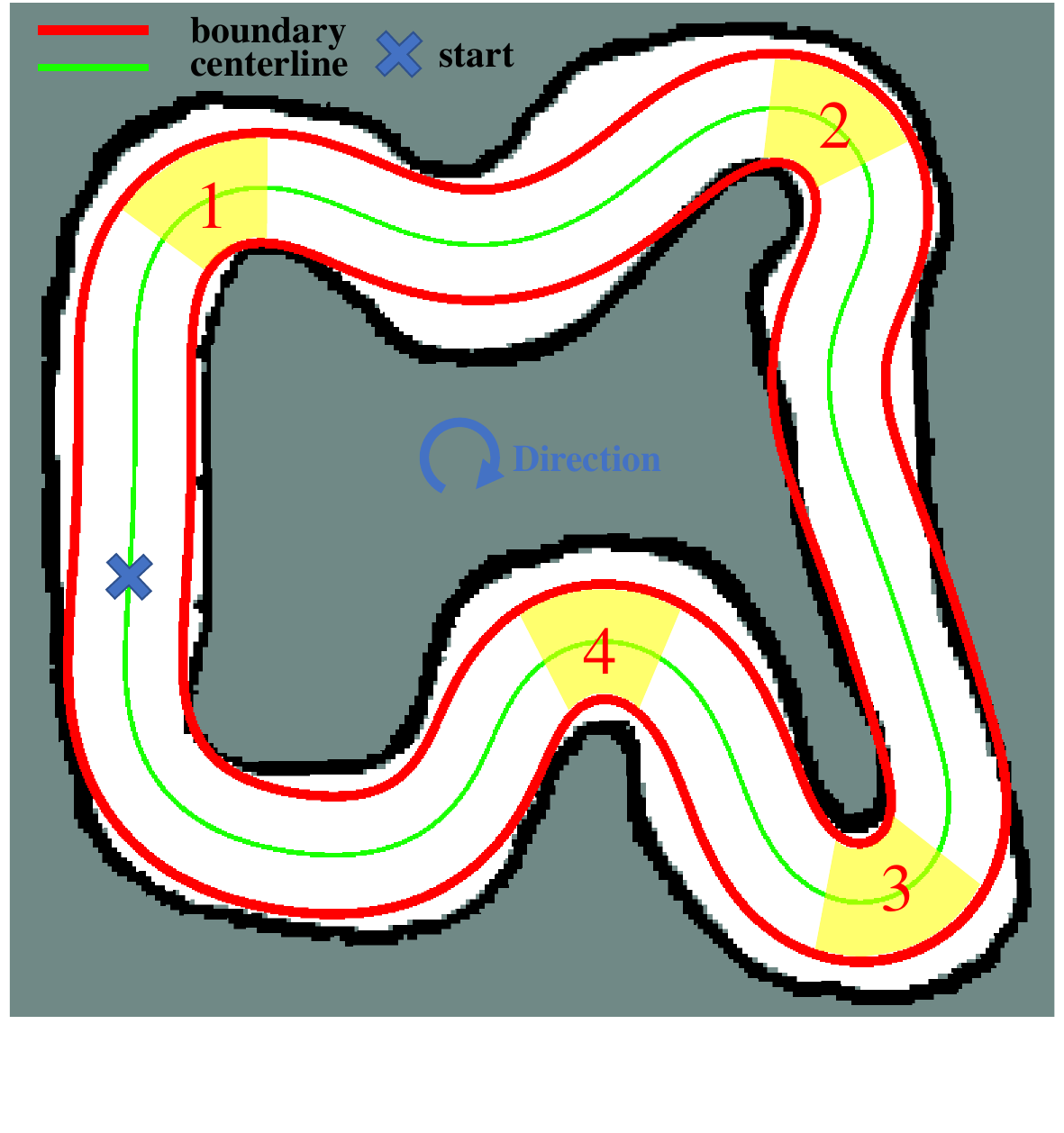}\label{fig:track_yaml}}
	\caption{The background color in the left figure corresponds to the NSC, corresponding to the point where the actual pose of the vehicle is projected onto the racetrack centerline. The numbers (1,2,3,4) in the left figure correspond to the four large curvature turns in the right figure. Compared to the MPCC method, CiMPCC significantly improves deceleration performance in sharp turns, allowing it to maintain high velocities on racetrack sections with minor curvatures.}
	\label{fig:v-samples}
\end{figure*}

We build a 1:10 scale autonomous racing vehicle based on the F1TENTH software system\cite{o2020f1tenth}, as shown in Fig.~\ref{fig:real_car}. This self-built 1:10 scale vehicle is designed for Driving, Drifting, and Racing Automation (DDRA). The  computational unit NUC runs ROS Melodic under Ubuntu 18.04 and the solver used for the optimization problem is CasADi\cite{Andersson2019}. We utilize IMU, 2D lidar, and odometer data to construct the map through Cartographer ROS\cite{45466}, and the localization is performed using particle filter\cite{8460743}. We simultaneously built an actual racetrack shown in Fig.~\ref{fig:racetrack}. In Fig.~\ref{fig:track_yaml}, the map after the completion of map construction using Cartographer ROS is presented through rviz visualization.

To begin with, we compare the CiMPCC with the original MPCC to verify the effectiveness of incorporating NSC into the objective function. 

\begin{table*}
	\begin{minipage}{0.70\linewidth}
		\centering
		\caption{Mean Lap Time and Velocity During seventeen Laps of Autonomous Racing for three Trajectory Planning Methods}\label{tab:data}
		\begin{tabular}{cp{3.665em}ccccc}
			\toprule
			& \multicolumn{1}{r}{} & \multicolumn{1}{p{4.4em}}{\textbf{MPCC}} & \multicolumn{1}{p{4.9em}}{\textbf{\fontsize{7}{1}\selectfont Comparison}} & \multicolumn{1}{p{4.8em}}{\textbf{RDM+OTG}} & \multicolumn{1}{p{4.9em}}{\textbf{\fontsize{7}{1}\selectfont Comparison}} & \multicolumn{1}{p{4.4em}}{\textbf{CiMPCC}} \\
			\midrule
			\midrule
			\multicolumn{1}{c}{\multirow{3}[2]{*}{\textbf{Lap Time in [s]}}} & \textbf{Max} & 16.654  & $\uparrow$12.5\% & 16.437  & $\uparrow$11.4\% & \textbf{14.568 }\\
			& \textbf{Min} & 15.801  & $\uparrow$11.7\% & 15.743  & $\uparrow$11.4\% & \textbf{13.945 }\\
			& \textbf{Mean} & \cellcolor[rgb]{ .906,  .902,  .902}16.106  & \cellcolor[rgb]{ .906,  .902,  .902}$\uparrow$11.8\% & \cellcolor[rgb]{ .906,  .902,  .902}16.083  & \cellcolor[rgb]{ .906,  .902,  .902}$\uparrow$11.7\% & \cellcolor[rgb]{ .906,  .902,  .902}\textbf{14.202 }\\
			\midrule
			\multicolumn{1}{c}{\multirow{3}[2]{*}{\textbf{Velocity in [m/s]}}} & \textbf{Max} & 2.963  & $\downarrow$ -14.9\% & 2.901  & $\downarrow$ -17.3\% & \textbf{3.404 } \\
			& \textbf{Min} & 2.816  & $\downarrow$ -17.0\% & 2.819  & $\downarrow$ -16.9\% & \textbf{3.294 } \\
			& \textbf{Mean}&\cellcolor[rgb]{ .906,  .902,  .902}2.910  & \cellcolor[rgb]{ .906,  .902,  .902}$\downarrow$ -15.2\% & \cellcolor[rgb]{ .906,  .902,  .902}2.864  & \cellcolor[rgb]{ .906,  .902,  .902}$\downarrow$ -17.0\%&\cellcolor[rgb]{ .906,  .902,  .902}\textbf{3.351}\\
			\bottomrule
		\end{tabular}%
	\end{minipage}%
	\hspace{\fill}%
	\begin{minipage}{0.3\linewidth}
		\centering
		\caption{Vehicle Parameters and \\ Algorithm Parameters}\label{tab:paras}
		\begin{tabular}{ll}
			\toprule
			\multicolumn{1}{p{4.235em}}{\textbf{Parameter}} & \multicolumn{1}{p{4.235em}}{\hspace{0.3cm}\textbf{Setting}} \\
			\midrule
			\midrule
			$L$   & 0.324m \\
			$N_p,N_c$ & 10 \\
			$u_{\textrm{min}}$ & [-10m/s -0.35rad-10m/s] \\
			$u_{\textrm{max}}$ & [10m/s 0.35rad 10m/s] \\
			$\bar{\mathbf{v}}$ &   [4.18m/s 3.8m/s]\\
			$\underline{\mathbf{v}}$ & [2.72m/s 2.47m/s] \\
			\bottomrule
		\end{tabular}%
	\end{minipage}
\end{table*}

The vehicle model used for MPCC and CiMPCC is:
\begin{equation}\label{eq:diff_eq}
	\dot{\zeta}=f(x,u) \triangleq \quad
	\begin{aligned}
		\dot{X} &= \cos (\varphi) \cdot \mathbf{v}_{l}  \\ 
		\dot{Y}&= \sin (\varphi) \cdot \mathbf{v}_{l} \\ 
		\dot{\varphi}&= \tan (\delta)/{L} \cdot \mathbf{v}_{l}  \\
		\dot{s}&= \mathbf{v}_{p}
	\end{aligned}
\end{equation}
where $\delta$ is the steering angle and $\varphi$ is the yaw angle. $L$ is the wheelbase. $X,Y$ are Cartesian coordinates of the vehicle rear axle center. $s$ is the distance the vehicle travels along the racetrack centerline. $\zeta={\begin{bmatrix} X & Y & \varphi & s \end{bmatrix}}$ is the state and $u={ \begin{bmatrix}  \mathbf{v}_{l} & \delta & \mathbf{v}_{p}  \end{bmatrix}}$ is the control variable.  The final control inputs for the underlying chassis are the vehicle body velocity $\mathbf{v}_{l}$ and the steering angle $ \delta $.

The multiple-shooting\cite{zhangGuaranteedCollisionFree2021} is introduced for improving the efficiency of solving the optimization problem~\eqref{cimpcc}. The physical parameters and constraints of the DDRA are detailed in Table~\ref{tab:paras}. 
\textcolor{black}{During runtime, the parameters of the optimization problem for MPCC and CiMPCC are finetuned to achieve the best performance. In the context of the CiMPCC and MPCC optimization problems, the diagonal matrices $Q$ and $R_1$, along with the scalar $\gamma$, share same values: $Q = \text{diag}([800, 800])$, $\gamma = 40$, $R_1 = \text{diag}([10, 3500, 0])$. In the traditional MPCC, the reference control variable $u_{\text{ref}}$ are set to $\text{diag}([3.3, 0, 3])$ and $R_2$ is $\text{diag}([40, 10, 40])$. However, for CiMPCC, which maps the reference velocity according to curvature, the reference velocity component from the $u_{\text{ref}}$ is omitted, effectively setting $R_2$ to $\text{diag}([0, 10, 0])$. Furthermore, for the CiMPCC, $R_3 = \text{diag}([40, 40])$.}

For the $\mathbf{v}={\begin{bmatrix} \mathbf{v}_l & \mathbf{v}_{p} \end{bmatrix}}$ which is OV setting, the aggressive projected velocity $\bar{\mathbf{v}}_p = 3.8 \textrm{m/s}$ is obtained as the fastest velocity at which the expert drives the vehicle to complete a single lap. The corresponding vehicle body velocity is $\bar{\mathbf{v}}_l = \bar{\mathbf{v}}_p \cdot 1.1$. Safe racing velocities $\underline{\mathbf{v}}$ are derived from the aggressive velocities with a discount factor of $0.65$. The specific settings are shown in Table~\ref{tab:paras}. 

The control inputs of MPCC and CiMPCC in a complete lap are shown in Fig~\ref{fig:samples}. \textcolor{black}{It is evident that CiMPCC maintains low velocity in sections of high curvature due to the presence of the LTC.  In contrast, compared to MPCC, CiMPCC ensures higher velocity in sections with small curvature owing to the presence of the UTC. \textcolor{black}{The reason MPCC fails to decelerate sufficiently in curved racetrack sections is its inability to account for the curvature of the centerline, leading to a failure to maintain velocity comparable to CiMPCC in sections with small curvature.}  Through the comparison and analysis of control inputs, it can be concluded that CiMPCC effectively models the NSC of the racetrack centerline.}

\newcommand{\imgscale}{0.22}
\begin{figure*}[!t]
	\centering
	\subfloat[The racing trajectory of the MPCC.]{\includegraphics[scale=\imgscale]{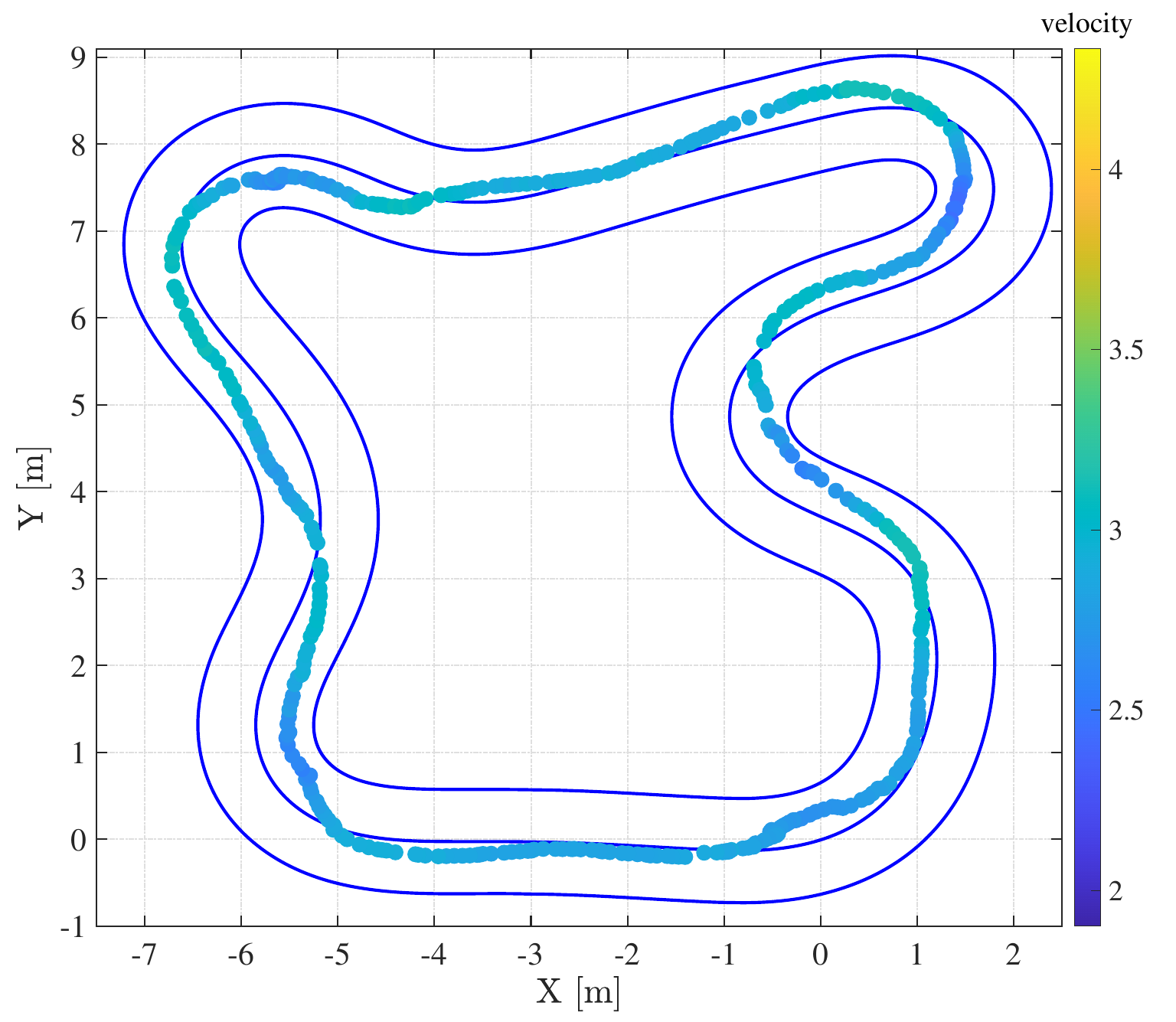}\label{fig:mpcc_race}}
	\hfil
	\subfloat[The racing trajectory of the CiMPCC.]{\includegraphics[scale=\imgscale]{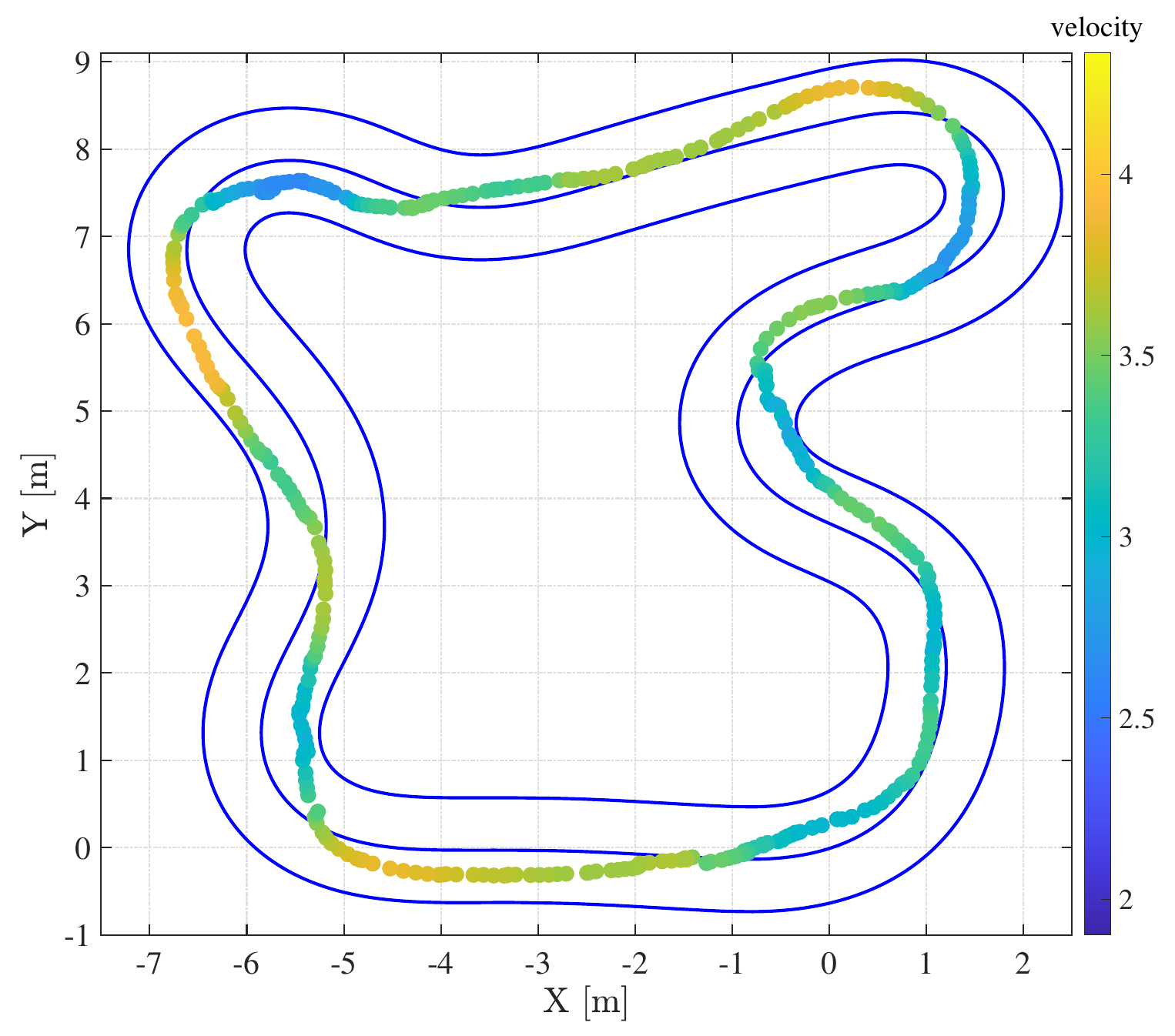}\label{fig:cimpcc_race}}
	\hfil
	\subfloat[The racing trajectory of the RMD+OTG.]{\includegraphics[scale=\imgscale]{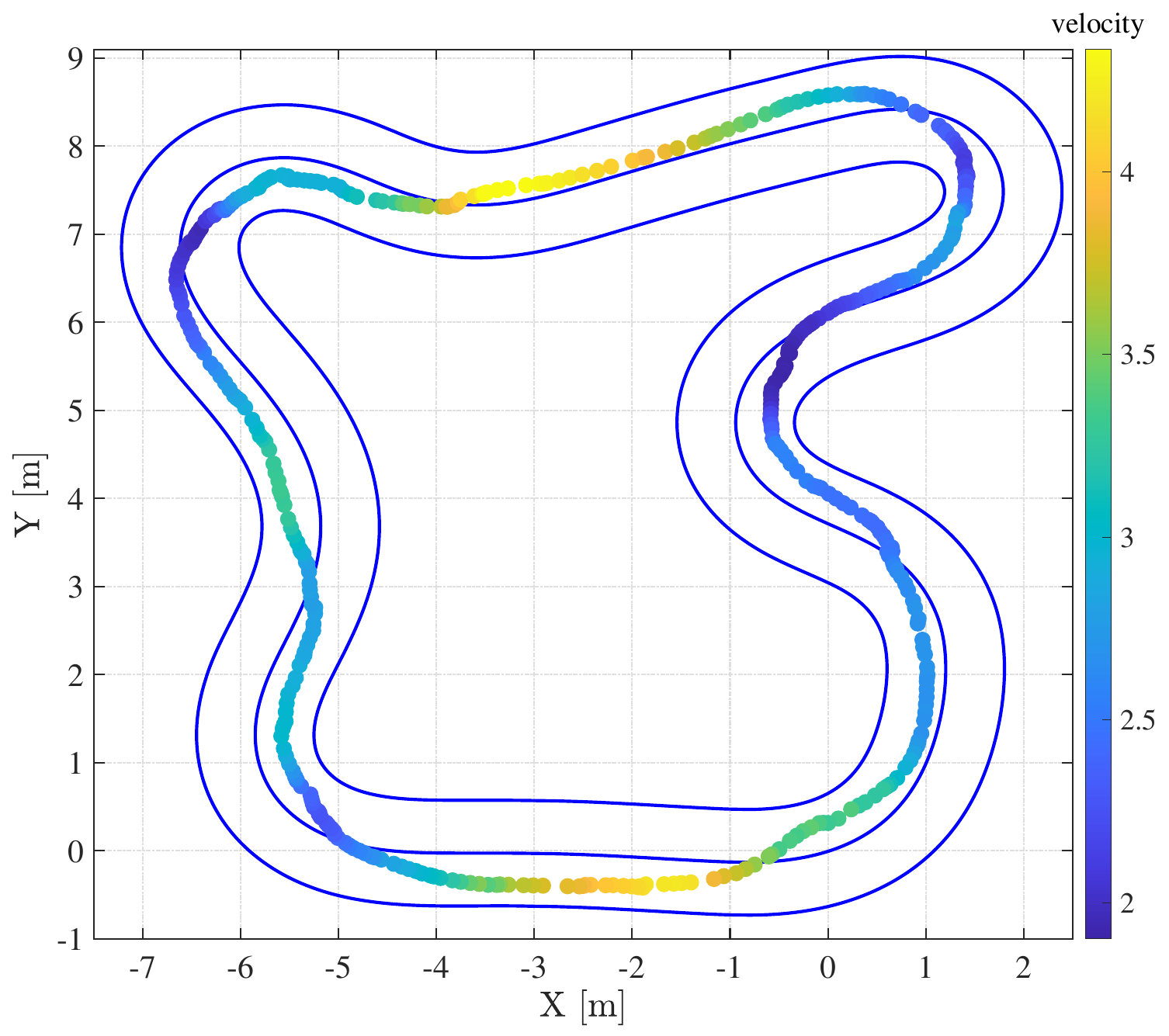}\label{fig:pp_race}}
	\caption{Results of three different trajectory planning methods applied to DDRA for autonomous racing. The trajectory points are obtained by Particle Filter.}
	\label{fig:race_traj}
\end{figure*}

To further demonstrate advantages of CiMPCC compared to global trajectory planning methods, we also introduce a comparative global trajectory planning method for autonomous racing based on path-velocity decomposition. Minimum curvature is used for path planning\cite{braghinRaceDriverModel2008} (referred to as RDM in the following), and velocity planning is used as in \cite{werlingOptimalTrajectoryGeneration2010} (referred to as OTG in the following).

The racing performance of CiMPCC is compared with MPCC and RDM+OTG methods in the following analysis. The trajectory planned by RDM+OTG is tracked by the proposed controller in \cite{becker2023model}. The Fig.~\ref{fig:race_traj} illustrates trajectories of autonomous racing by applying these three methods on the DDRA vehicle. The CiMPCC method exhibits higher velocity than the MPCC method during racing. Compared with the RDM+OTG method, the CiMPCC method has a significantly higher velocity at corner entry and exit, which is due to the presence of the LTC that allows the CiMPCC to decelerate quickly in corners.

Subsequently, we continuously conduct the DDRA for a total of seventeen laps of autonomous racing. The velocity, lap time and computation time of CiMPCC are recorded while applying these methods. 

\begin{figure}[!t]
	\centering 
	\includegraphics[scale=0.19]{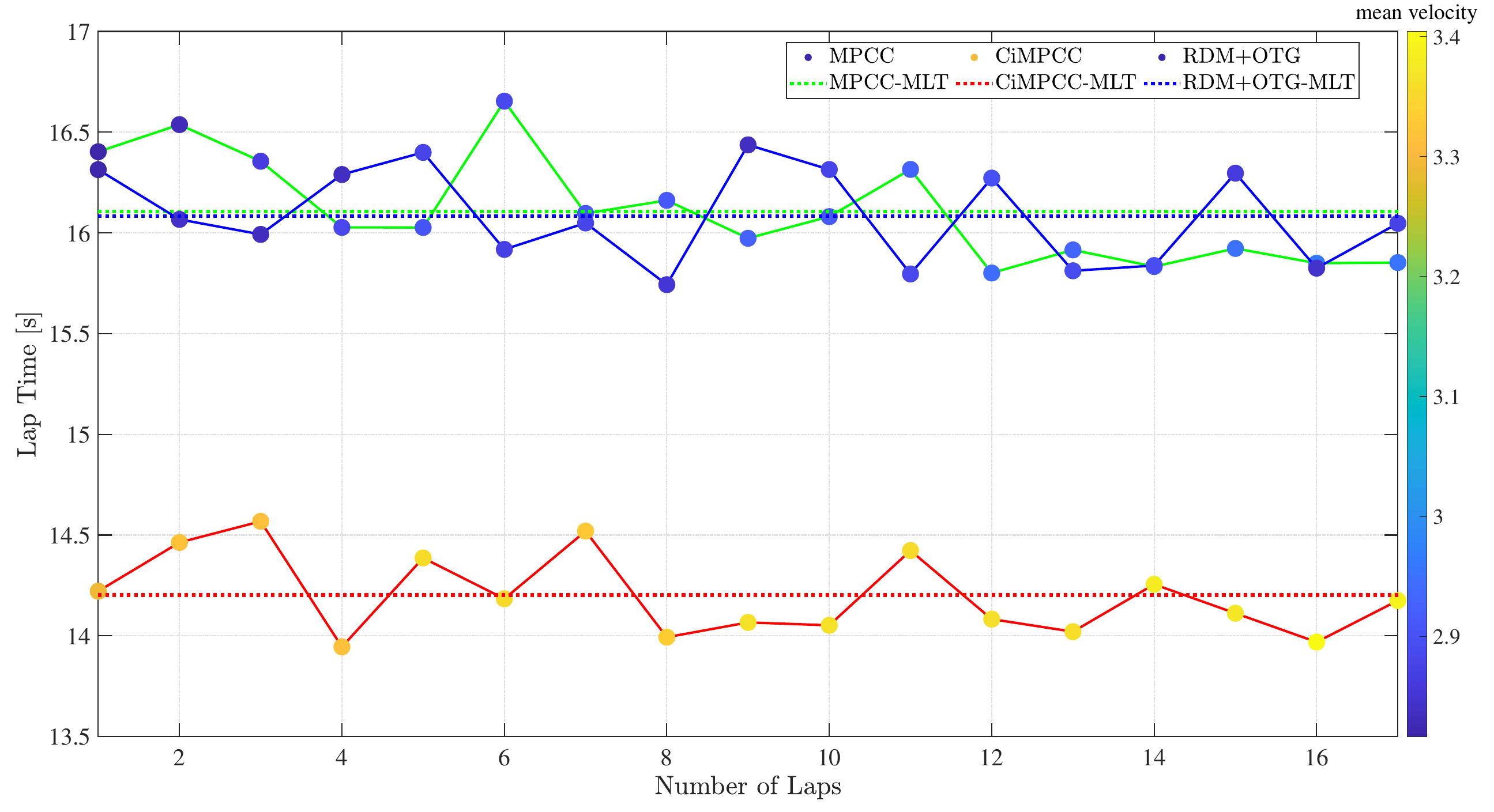}
	\caption{Mean velocity of a single lap, lap time, and mean lap time(MLT) in 17-lap autonomous racing for three trajectory planning methods.}
	\label{fig:mean_v}
\end{figure}

The statistics of the mean velocity and lap time are shown in Table~\ref{tab:data}.  It can be noted that MLT of CiMPCC is decreased by $\textbf{11.8\%}$ compared to MPCC and by $\textbf{11.7\%}$ compared to RDM+OTG. The mean velocity is increased by $\textbf{15.2\%}$ compared to MPCC and $\textbf{17.2\%}$ compared to RDM+OTG. According to the percentage of performance improvement, the CiMPCC method outperforms the traditional MPCC and RDM+OTG methods in terms of velocity and lap time. The schematic diagram of lap time, mean velocity of a single lap, and mean lap time (MLT) of seventeen laps during continuous autonomous racing is shown in Fig.~\ref{fig:mean_v}.

\begin{figure}[!t]
	\centering 
	\includegraphics[scale=0.19]{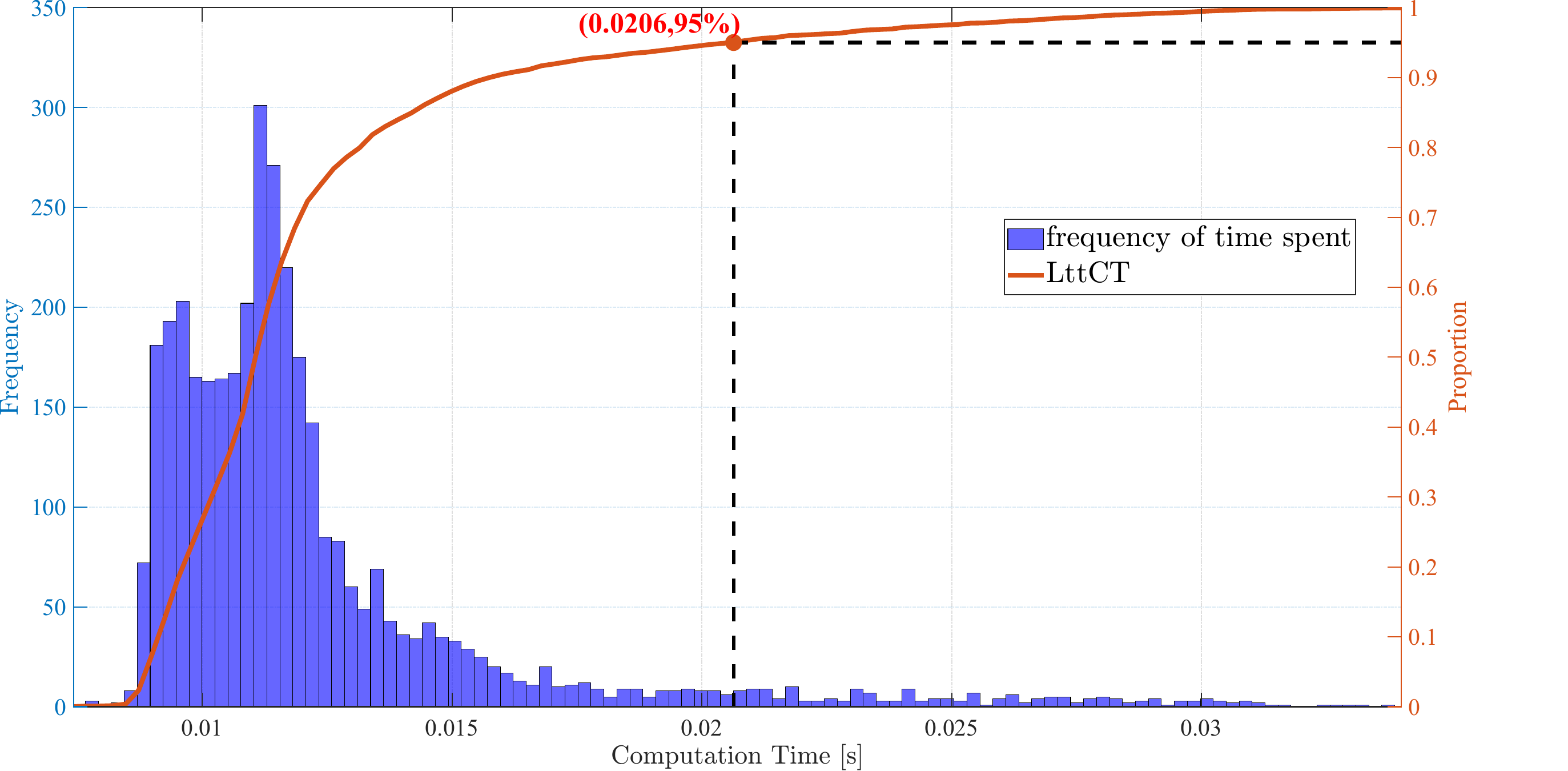}
	\caption{Histogram of CiMPCC computation time. The brown curve represents the proportion of samples for which the computation time is less than the corresponding time (LttCT), out of the total sample size.}
	\label{fig:time}
\end{figure}

The statistical data on computation time of CiMPCC are shown in Fig.~\ref{fig:time}. Since CiMPCC is a nonlinear programming problem, we use the result of the previous instant as the warm start for the current instant. During the seventeen-lap run, over \textbf{95\%} of CiMPCC computation times are less than \textbf{0.0206} seconds. This demonstrates that the computation efficiency of CiMPCC is sufficient for real-time computation.

The experimental part is summarized. Firstly, the effectiveness of incorporating the curvature of the racetrack centerline into the CiMPCC optimization problem to optimize velocity is verified. This is verified by comparing the control inputs of CiMPCC and MPCC. Secondly, CiMPCC significantly improves lap time and velocity in a long-time autonomous racing task compared to MPCC and RMD+OTG methods. In addition, the computation efficiency of the CiMPCC method is sufficient to meet the real-time requirements.

\section{Conclusion}\label{Conclusion}

In this paper, we introduce a novel curvature-integrated local trajectory planning method, known as CiMPCC, for autonomous racing. The key innovation of this method is the integration of the racetrack centerline into the optimization problem. CiMPCC employs a nonlinear continuous function to establish a curvature-velocity relationship, optimizing the velocity in the planned local trajectory for autonomous racing. The proposed method is validated using a self-built 1:10 scale DDRA vehicle. Experimental results from long-term autonomous racing show that the CiMPCC method decreases the mean lap time by \textbf{11.8\%} compared to the traditional MPCC method, confirming its effectiveness. Additionally, it demonstrates the feasibility of incorporating the curvature of the racetrack centerline into the optimization problem to reduce lap time in autonomous racing.

In our future work, we plan to implement the proposed method in more complex scenarios such as overtaking. Additionally, we will evaluate the impact of vehicle model mismatch on the CiMPCC during autonomous racing.

\section*{Acknowledgments}
The authors would like to thank Yiqin Wang, Haoyang He, and Hongxu Chen for their help with the perception system during the construction of the DDRA vehicle system. Thanks also to Wangjia Weng, Wule Mao, and Jihao Huang for their valuable insights on autonomous racing. Thanks to Bei Zhou for her insightful suggestions on the structure of the paper. We thank all reviewers and editors for their precious and insightful comments.

\bibliographystyle{IEEEtran}
\bibliography{bibref_zotero, bibref.bib}

\end{document}